\documentclass[wcp]{jmlr}

\usepackage{longtable}
\usepackage{booktabs}
\usepackage{amssymb}
\usepackage{latexsym}
\usepackage[utf8]{inputenc} 
\usepackage[T1]{fontenc}    
\usepackage{hyperref}       
\usepackage{url}            
\usepackage{booktabs}       
\usepackage{amsfonts}       
\usepackage{nicefrac}       
\usepackage{microtype}      
\usepackage{bm}             
\usepackage{graphicx}       
\usepackage{algorithm}      
\usepackage[noend]{algpseudocode} 
\usepackage{caption}        
\usepackage{array}          
\usepackage{booktabs}       
\usepackage{pbox}           
\usepackage{subcaption} 

\hypersetup{
    colorlinks = true,
}
\makeatletter
\def\BState{\State\hskip-\ALG@thistlm}
\makeatother
\floatname{algorithm}{Procedure}

\title{Grouping Capsules Based Different Types}

\author{\Name{Qiang Ren} \Email{rqfzpy@tongji.edu.cn}\\
\addr Department of Computer Science and Technology, Tongji University, Shanghai 201804, China.}

\begin{document}

\maketitle

\begin{abstract}
Capsule network was introduced as a new architecture of neural networks, it encoding features as capsules to overcome the lacking of equivariant in the convolutional neural networks. It uses dynamic routing algorithm to train parameters in different capsule layers, but the dynamic routing algorithm need to be improved. In this paper, we propose a novel capsule network architecture and discussed the effect of initialization method of the coupling coefficient $c_{ij}$ on the model. First, we analyze the rate of change of the initial value of $c_{ij}$ when the dynamic routing algorithm iterates. The larger the initial value of $c_{ij}$, the better effect of the model. Then, we proposed improvement that training different types of capsules by grouping capsules based different types. And this improvement can adjust the initial value of $c_{ij}$ to make it more suitable. We experimented with our improvements on some computer vision datasets and achieved better results than the original capsule network.
\end{abstract}

\section{Introduction}

In recent years, the rise of deep learning has brought great success in the research of computer vision, and has created outstanding achievements in computer vision such as image classification\citep{krizhevsky2012imagenet}; \cite{simonyan2014very}; \cite{szegedy2015going}; \cite{zeiler2014visualizing}; \citep{he2016deep}, image detection\citep{girshick2015fast}; \citep{ren2015faster}; \citep{he2017mask}; \citep{szegedy2013deep}; \citep{girshick2014rich}; \citep{redmon2016you}; \citep{liu2016ssd} and image segmentation\citep{long2015fully}; \citep{ronneberger2015u}. especially convolutional neural networks (CNN) have achieved remarkable progress on computer vision tasks. Although the stacking of convolutions filters and non-linearity units can make the network have better learning ability, CNN is not essentially the viewpoint invariant. This means that the spatial relationship between different features is usually not learned when using CNN.

In order to address these limitations, Hitton and Sabour proposed capsule networks\citep{sabour2017dynamic}.The capsule networks enhances its representative ability by encapsulating information in an activation vectors called a capsule. The scalar output neurons of traditional neural networks are replaced by vector output capsules in the capsule network, which are collections of neurons. A capsule represents an object, or part of an object, and the activity vector of the capsule encodes the instantiation parameters of this part.

The proposed model that achieves state-of-the-art performance on MNIST\citep{lecun1998gradient} and is better than convolutional networks in identifying highly overlapping numbers. The capsule network uses dynamic routing algorithm to train the network between capsule layers. The two core parameters of the dynamic routing algorithm are weight matrix $W_{ij}$ and coupling coefficients $c_{ij}$. The matrix $W_{ij}$ is an affine transformation matrix for learning the transformation of objects in different directions. The coupling coefficients $c_{ij}$ is the connection weight coefficient between the capsule of the layer below and the capsule of the layer above.

In the original capsule networks, multiply all the capsules on the layer below and coupling coefficients $c_{ij}$ to get a capsule on the layer above. Such an algorithm has some disadvantages. First, in the original model, the capsules are divided into 32 groups, because the convolution kernels used to extract features are different, so the type of each group can be regarded as different.Second, some types of capsules on the layer below may have a less contribution to the classification task, and all of the layer below participating in the calculation may have some effect on the task. Finally, all the capsule on the layer below are involved in the calculation, and different types of capsules may have some conflicts or other effects. Therefore, I proposed to make different types of capsules use dynamic routing algorithms, and then classify the object according to the obtained capsules on the layer above.

To this end, in this paper, we explore the architecture of the original model, and proposed a variant of the capsule networks, called the Grouping Capsules Networks(G-CapsNet). We propose the new architecture of capsule networks achieved different types of capsules are grouped for training, and propose a discussion on the method of coupling coefficients $c_{ij}$ initialization. More specifically, we make the following contributions in the paper:
\begin{enumerate}
\item We proposed different initialization methods for the coupling coefficients $c_{ij}$ and different initialization methods have different effects on the model.
\item We discussed the reasons for the results in $1$.
\item We proposed a novel capsule networks architecture called the Grouping Capsules Networks(G-CapsNet). This new capsule networks architecture allows different types of capsules are grouped for training and reduce the effect of coupling coefficient $c_{ij}$ initialization methods on the results.
\end{enumerate}
\section{Related work}
\subsection{Capsule Networks}

The capsule networks is a new neural network model proposed by Hitton, which aims to solve some shortcomings of convolutional neural networks. The idea of capsule originated with a paper published by Hitton in 2011, it showed how to use neural networks to learn the characteristics of the entire vector of output instantiation parameters\citep{hinton2011transforming}. However, it did not cause a big trend, and it only entered the public's field of vision in 2017. In 2017, Hitton and Sabour proposed the original structure of the capsule networks, which uses dynamic routing algorithms to train parameters between capsule layers\citep{sabour2017dynamic} They want the length of the output vector of the capsule to represent the probability of the entity being present. In order to make the capsules more nonlinear, a non-linear {\bf squashing} function is used to ensure that the shorter capsules are shrunk to a length of almost zero and the longer capsules are shrunk to a length slightly below one.
\begin{equation}
{\bf v}_j = \frac{||{\bf s}_j||^2}{1+||{\bf s}_j||^2} \frac{{\bf s}_j}{||{\bf s}_j||}
\label{squash}
\end{equation}

Where the ${\bf v}_j$ in the Equation \ref{squash} is the vector output of capsule $j$, and ${\mathbf{s}}_j$ is the weighted sum of all of its corresponding input capsules.
The total input of capsule ${\mathbf{s}}_j$ is the weighted sum of all ``prediction vectors'' ${\bf \hat{u}}_{j|i}$ from the capsules in the lower layer, and is produced by multiplying the output  ${\bf u}_i$ of the capsules by a weight matrix  ${\bf W}_{ij}$.
\begin{equation}
{\bf s}_j = \sum_i c_{ij} {\bf \hat{u}}_{j|i} \ , \ \ \ \ \ \ \ 
{\bf \hat{u}}_{j|i} = {\bf W}_{ij}{\bf u}_i 
\label{equation2}
\end{equation}

Where the $c_{ij}$ in the Equation \ref{equation2} is the coupling coefficient determined by the dynamic routing algorithm.
The coupling coefficients $c_{ij}$ is capsule $i$ for all the capsules in the upper layer sum to $1$ and are determined by a ``routing softmax'' whose initial logits priors $b_{ij}$ that capsule~$i$ should be coupled to capsule~$j$. 
\begin{equation}
c_{ij} = \frac{\exp(b_{ij})}{\sum_k \exp(b_{ik})}
\label{softmax}
\end{equation}

The logits priors $b_{ij}$ uses dynamic routing algorithm to learn in the process of capsule network training and then iteratively refined by measuring the agreement between the current output ${\bf v}_j$ of each capsule~$j$  in the upper layer and the prediction ${\bf \hat{u}}_{j|i}$ made by capsule~$i$.

In convolutional capsule layers, each capsule outputs a local grid of vectors to each type of capsule in the layer above using different transformation matrices for each member of the grid as well as for each type of capsule.

\begin{algorithm}[h]

	\caption{Algorithm1.}
	\label{routingalg1}
	\KwIn{$\bm{\hat{u}}_{j|i}$, $r$, $l$}
	\KwOut{${\bf v}_j$}
	for all capsule $i$ in layer $l$ and capsule $j$ in layer $(l+1)$: $b_{ij} \gets 0$.

	\For{$r$ iterations}{
for all capsule $i$ in layer $l$: ${\bf c}_i \gets \texttt{softmax}({\bf b}_i)$

for all capsule $j$ in layer $(l+1)$: ${\bf s}_j \gets \sum_i{c_{ij}{\bf \hat u}_{j|i}}$

for all capsule $j$ in layer $(l+1)$: ${\bf v}_j \gets \texttt{squash}({\bf s}_j)$

for all capsule $i$ in layer $l$ and capsule $j$ in layer $(l+1)$: $b_{ij} \gets b_{ij} + {\bf \hat{u}}_{j|i} . {\bf v}_j$	
}
\end{algorithm}

The process of original dynamic routing can be described as {\bf Algorithm \ref{routingalg1}}.

\subsection{Related Work}
After the capsule network was proposed based on the dynamic routing algorithm, various improvements of dynamic routing algorithm appeared. Hinton then proposed a new method for training capsule networks. They discussed matrix capsules and applies EM (expectation maximization) routing to classify images at different angles\citep{sabour2018matrix}. After that, the calculation of the amount of the capsule network has become a research direction of the capsule networks. Wang improved the calculation method of dynamic routing pair, proposed a new dynamic routing and other calculation methods, improved the initialization and update strategy of the coupling coefficient, made the dynamic routing algorithm converge faster, and in a simple unsupervised Experiments were carried out in the task of learning, and a better experimental result was obtained than the capsule of the initial version\citep{wang2018an}. Shahroudnejad proposed an analysis of the structure and effect of the capsule networks, and gave the potential interpretability of the capsule networks, and showed that the capsules were combined in different structures, giving the capsule networks a better explanation\citep{shahroudnejad2018improved}. Rawlinson proposed that the supervised capsule network could not be stacked very deeply, which would result in classification equivalence. To solve this problem, a sparse capsule network was proposed and a sparse unsupervised capsule network was trained to classify the experimental SVM layer. The effect on affNIST has been greatly improved\citep{rawlinson2018sparse}. Zhang proposed to improve the existing dynamic routing algorithm under the idea of weighted kernel density, and proposed two different optimization strategies for fast routing methods, which improved the time efficiency by 40\%, but performance has not decreased\citep{zhang2018fast}. Lu proposed an Affine Transformation Capsule Net (AT-CapsNet) which leverage both of the length and orientation information of digit capsules by adding a single-layer perceptron substitutes for the operation of computing length of vectors\citep{lu2018affine}. Li introduced a psychological theory which is called Cognitive Consistency to optimize the routing algorithm of Capsule Networks to make it more close to the working pattern of the human brain\citep{li2019cognitive}. Ferrarini aims to fill this shortcoming that Capsule Networks has low tolerance to imbalanced data and proposes two experimental scenarios to assess the tolerance to imbalanced training data and to determine the generalization performance of a model with unfamiliar affine transformations of the images\citep{ferrarini2019assessing}. Peer added a bias parameter to the routing-by-agreement algorithm. And they proved that without such a term the representation of activation vectors is limited and this becomes a problem for deeper capsule networks\citep{peer2019limitations}. 

With the development of capsule networks, many improved capsule networks models have been proposed. Deli{\`e}ge proposed the HitNet that a neural network with capsules embedded in a Hit-or-Miss layer, extended with hybrid data augmentation and ghost capsules. They tried to train to hit or miss a central capsule by tailoring a specific centripetal loss function and show how their network is capable of synthesizing a representative sample of the images of a given class by including a reconstruction network\citep{deliege2018hitnet}. Rosario proposed the Multi-Lane Capsule Networks (MLCN), which are a separable and resource efficient organization of Capsule Networks. It is composed of a number of (distinct) parallel lanes, each contributing to a dimension of the result, trained using the routing-by-agreement organization of Capsule Networks\citep{rosario2019multi}. Amer introduce a deep parallel multi-path version of Capsule Networks called Path Capsule Networks. It show that a judicious coordination of depth, max-pooling, regularization by DropCircuit and a new fan-in routing by agreement technique can achieve better or comparable results to Capsule Networks, while further reducing the parameter count significantly\citep{Amer2019Path}. Peer introduce a new routing algorithm called dynamic deep routing. It overcomes the routing-by-agreement algorithm does not ensure the emergence of a parse tree in the network, allows the training of deeper capsule networks and is also more robust to white box adversarial attacks than the original routing algorithm\citep{peer2018training}. Ribeiro present group equivariant capsule networks that a framework to introduce guaranteed equivariance and invariance properties to the capsule networks. They present a generic routing by agreement algorithm defined on elements of a group and prove that equivariance of output pose vectors, as well as invariance of output activations, hold under certain conditions. They present a new capsule routing algorithm based of Variational Bayes for a mixture of transforming gaussians. The Bayesian approach addresses some of the inherent weaknesses of EM routing such as the 'variance collapse' by modelling uncertainty over the capsule parameters in addition to the routing assignment posterior probabilities\citep{ribeiro2019capsule}. Phaye proposed frameworks customize the Capsule Networks by replacing the standard convolutional layers with densely connected convolutions. This helps in incorporating feature maps learned by different layers in forming the primary capsules. This improvement has been added a deeper convolution network which leads to learning of discriminative feature maps and uses a hierarchical architecture to learn capsules that represent spatial information which makes it more efficient for learning complex data\citep{phaye2018dense}. Rajasegaran introduced DeepCaps, a deep capsule networks architecture which uses a novel 3D convolution based dynamic routing algorithm. And they proposed a class-independent decoder network, which strengthens the use of reconstruction loss as a regularization term\citep{rajasegaran2019deepcap}.
\section{Methodology}
\subsection{Problem Formulation}
In the original capsule networks, the $c_{ij}$ is called the coupling coefficient and the $b_{ij}$ are the logits priors. All $b_{ij}$ is initialized to 0, $c_{ij}$ is the softmax of all the capsules in the layer above calculated by $b_{ij}$, so the probability of capsules in the layer above being activated at the beginning is equal. Since the category of the capsules in the layer above is $10$, each $c_{ij}$ is initialized to $0.1$.

\begin{equation}
c_{ij} = \frac{\exp(b_{ij})}{\sum_k \exp(b_{kj})}
\label{softmax1}
\end{equation}
The new initialization of $c_{ij}$ is Equation \ref{softmax1}.

\begin{algorithm}[h]
	\caption{Algorithm2.}
	\label{routingalg2}
	\KwIn{$\bm{\hat{u}}_{j|i}$, $r$, $l$}
	\KwOut{${\bf v}_j$}
	for all capsule $i$ in layer $l$ and capsule $j$ in layer $(l+1)$: $b_{ij} \gets 0$.

	\For{$r$ iterations}{
for all capsule $i$ in layer $l$: ${\bf c}_j \gets \texttt{softmax}({\bf b}_j)$

for all capsule $j$ in layer $(l+1)$: ${\bf s}_j \gets \sum_i{c_{ij}{\bf \hat u}_{j|i}}$

for all capsule $j$ in layer $(l+1)$: ${\bf v}_j \gets \texttt{squash}({\bf s}_j)$

for all capsule $i$ in layer $l$ and capsule $j$ in layer $(l+1)$: $b_{ij} \gets b_{ij} + {\bf \hat{u}}_{j|i} . {\bf v}_j$	
}
\end{algorithm}

The process of the new initialization of $c_{ij}$ can be described as {\bf Algorithm \ref{routingalg2}}.

From the Eq.~\ref{equation2}, ${\mathbf{s}}_j$ is the weighted sum of all of its corresponding input capsules in the lower layer. If $c_{ij}$ is the softmax of all the capsules in the layer below calculated by $b_{ij}$, This makes it logically more reasonable. Because the number of capsules in the lower layer is $1152$ in the original capsule networks, each $c_{ij}$ is initialized to approximately $0.0008$. After the capsule network is trained, different $c_{ij}$ initialization methods make the experimental results different. The experimental results are shown in {\bf Table \ref{tab1}}.

\begin{table*}[h]
\caption{\label{tab1}Experimental results of Algorithm\ref{routingalg1} and Algorithm\ref{routingalg2}}
\centering
\begin{tabular}{|l|l|l|l|l|l|l|l|}
\hline
Model & MNIST & F-MNIST & K-MNIST & SVHN & CIFAR-10 & SMALLNORB & AFFNIST\\
\hline
CapsNet & 99.65 & 93.02 & 98.30 & 93.65 & 76.05 & 89.70 &99.66\\
\hline
CapsNet-c & 99.61 & 92.45 & 97.44 & 91.94 & 72.82 & 82.71 & 99.51\\
\hline
\end{tabular}
\end{table*}

We tested our experiments in seven datasets(MNIST\citep{lecun1998gradient}, F-MNIST\citep{xiao2017fashion}, K-MNIST\citep{clanuwat2018deep}, SVHN\citep{svhn}, CIFAR-10(\cite{krizhevsky2009learning}), SMALLNORB\citep{huang2005smallnorb} and AFFNIST(\cite{tieleman2013affnist})) and the experiments results are shown in {\bf Table \ref{tab1}}. From the {\bf Table \ref{tab1}}, We have gotten a significant drop in the experimental results of the seven image datasets if the initialization of $c_{ij}$ is changed. First, I think this result is because the initialization of $c_{ij}$ is different. The $c_{ij}$ is initialized to $0.1$ in the original capsule networks, but $c_{ij}$ is initialized to $0.0008$ in the our capsule networks. The value of $c_{ij}$ initialization in our method is too small, and the dynamic routing algorithm is an iterative calculation process. If the base of a variable is large, it will change even more faster during the iteration. The initial value of $c_{ij}$ in the original networks is larger than our method, so it changes faster during the iteration. The experimental results of the original capsule networks are better under the same number of iterations. 

Through the experimental results, I propose that in the appropriate range, under the same network configuration, the larger the value of $c_{ij}$ initialization, the better the experimental results.

\subsection{Capsule grouping}
\begin{figure}[H]
\begin{center}
\includegraphics[width=0.65\textwidth]{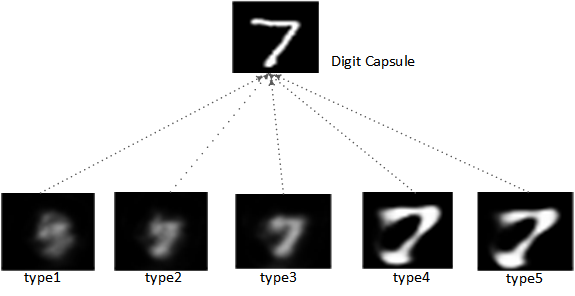}
\caption{Combination of different types of capsules}\label{p2}
\end{center}
\end{figure}

In the original capsule networks, there are $32$ different types of capsules on the PrimaryCapsules layer. Different types of capsules are characterized by different convolution kernels, and the same type of capsules are extracted from the same convolution kernels. Because each capsule is different in type, some capsules may be useless (or less contributing) to the classification. In the Figure \ref{p2}, the number represented by the digit capsule is $7$. It is a combination of many different types of capsules, but some capsules look similar to it, others are quite different. The type of capsule to which the similarity is larger, the greater the contribution to the judgment of the digit capsule category. Therefore, different types of capsules are grouped and trained, so that the capsule type with greater similarity to capsule j is trained, and its capsule is closer to capsule j after training. We train different types of capsules in groups so that the capsule type that is more similar to digit capsule is trained to be closer to the digit capsule.

\begin{figure}[htb]
\begin{center}
\includegraphics[width=1\textwidth]{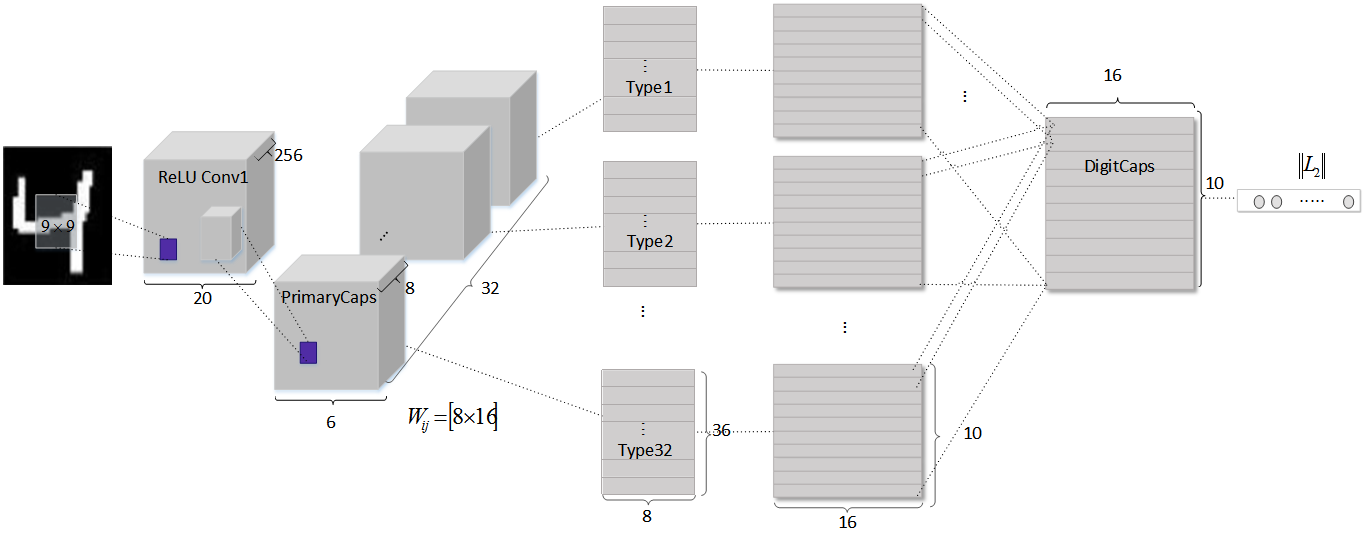}
\caption{Grouping Capsule Network Architecture}\label{p1}
\end{center}
\end{figure}

Therefore, we propose a new capsule network architecture in the Figure \ref{p1}, in which each type of capsule is trained using a dynamic routing algorithm, so that there are only $36$ capsules of each type. When using our method to initialize $c_{ij}$, it is initialized to approximately $0.028$, and it approximately $34$ times larger than the original network. The larger the base, the faster the update will be during the iteration, and the faster the update rate is the better result of the experiment. Similar to {\bf bagging} in ensemble Learning\citep{breiman1996bagging}, different types of capsule grouping training make the network model more robust and enhance network generalization ability. 

In the Figure \ref{p1}, our proposed capsule network architecture is similar to the original capsule network structure in the front and the end. The most significant difference is in the dynamic routing algorithm. In our networks architecture, after the PrimaryCaps layer, we group capsules by different capsule types. Then train them separately using dynamic routing algorithms, each type will get a digit capsule which represents the activation probability of different types of digit capsules. Finally, different types of digit capsules are combined into a total digit capsule and calculate its length.

\begin{algorithm}[h]
	\caption{Algorithm3.}
	\label{routingalg3}
	\KwIn{$\bm{\hat{u}}_{j|i}$, $r$, $l$}
	\KwOut{${\bf v}_j$}
	for all capsule $i$ in layer $l$ and capsule $j$ in layer $(l+1)$: $b_{ij} \gets 0$.

	\For{$r$ iterations}{
	\For{$m$ types}
	{
for all capsule $i$ of type $m$ in layer $l$: ${\bf c}_i \gets \texttt{softmax}({\bf b}_i)$

for all capsule $j$ of type $m$ in layer $(l+1)$: ${\bf s}_{jm} \gets \sum_i{c_{ij}{\bf \hat u}_{j|i}}$

for all capsule $j$ of type $m$ in layer $(l+1)$: ${\bf v}_{jm} \gets \texttt{squash}({\bf s}_{jm})$

}
${\bf v}_{j} \gets \texttt{squash}(\sum_m{\bf v}_{jm})$

for all capsule $i$ of in layer $l$ and capsule $j$ in layer $(l+1)$: $b_{ij} \gets b_{ij} + {\bf \hat{u}}_{j|i} . {\bf v}_j$	
}
\end{algorithm}

\begin{algorithm}[h]
	\caption{Algorithm4.}
	\label{routingalg4}
	\KwIn{$\bm{\hat{u}}_{j|i}$, $r$, $l$}
	\KwOut{${\bf v}_j$}
	for all capsule $i$ in layer $l$ and capsule $j$ in layer $(l+1)$: $b_{ij} \gets 0$.

	\For{$r$ iterations}{
	\For{$m$ types}
	{
for all capsule $i$ of type $m$ in layer $l$: ${\bf c}_j \gets \texttt{softmax}({\bf b}_j)$

for all capsule $j$ of type $m$ in layer $(l+1)$: ${\bf s}_{jm} \gets \sum_i{c_{ij}{\bf \hat u}_{j|i}}$

for all capsule $j$ of type $m$ in layer $(l+1)$: ${\bf v}_{jm} \gets \texttt{squash}({\bf s}_{jm})$

}
${\bf v}_{j} \gets \texttt{squash}(\sum_m{\bf v}_{jm})$

for all capsule $i$ of in layer $l$ and capsule $j$ in layer $(l+1)$: $b_{ij} \gets b_{ij} + {\bf \hat{u}}_{j|i} . {\bf v}_j$	
}
\end{algorithm}

The process of our capsule networks architecture can be described as {\bf Algorithm \ref{routingalg3}} and {\bf Algorithm \ref{routingalg4}}.(Two different methods of initializing $c_{ij}$)

\section{Experiment}
\subsection{Datasets and Implementation}
We test our new capsule networks architecture with several commonly used datasets in computer vision(MNIST\citep{lecun1998gradient}, F-MNIST\citep{xiao2017fashion}, K-MNIST\citep{clanuwat2018deep}, SVHN\citep{svhn}, CIFAR-10(\cite{krizhevsky2009learning}), SMALLNORB\citep{huang2005smallnorb} and AFFNIST(\cite{tieleman2013affnist})) and compare its performance with the original capsule network architecture. For CIFAR-10, SMALLNORB and SVHN, we resized the images to {$32 \times 32 \times 3$} and shifted by up to $2$ pixels in each direction with zero padding and no other data augmentation/deformation is used. For AFFNIST, We trained on MNIST with digit placed randomly on a black background of $40 \times 40$ pixels and tested this network on the AFFNIST. And for other datasets, original image sizes are used throughout our experiments.

We used pytorch libraries for the development of Experiment. For the training procedure, we used Adam optimizer with an initial learning rate of 0.001, which is reduced 5\% after each epochs\citep{Kingma2014Adam}. We set the batchsize is 128 that train with 128 images each time. The models were trained on GTX-1080Ti and training 100 epoch for every Experiment. All experiments were run three times and the results were averaged. 

\subsection{Experimental result}
For the original $c_{ij}$ initialization method(Initialize $c_{ij}$ to $0.1$), the experimental results of the new capsule network(Algorithm\ref{routingalg3}) is OurNet and the original capsule networks(Algorithm\ref{routingalg1}) is CapsNet are shown in {\bf Table \ref{tab2}}. 
\begin{table*}[h]
\caption{\label{tab2}Experimental results of Algorithm\ref{routingalg1} and Algorithm\ref{routingalg3}}
\centering
\begin{tabular}{|l|l|l|l|l|l|l|l|}
\hline
Model & MNIST & F-MNIST & K-MNIST & SVHN & CIFAR-10 & SMALLNORB & AFFNIST\\
\hline
CapsNet & 99.65 & 93.02 & 98.30 & 93.65 & 76.05 &  89.70 & 99.66\\
\hline
OurNet & \bf 99.68 & \bf 93.21 & \bf 98.33 & \bf 93.86 & \bf 76.25 & 89.17& 99.61\\
\hline
\end{tabular}
\end{table*}

 From the {\bf Table \ref{tab2}}, On the five datasets(MNIST, F-MNIST, K-MNIST, SVHN and CIFAR-10), after grouping different types of capsules for training, the experimental results are significantly improved compared with the original capsule networks. For the datesets of AFFNIST and SMALLNORB, the experimental results have declined. It is considered that after the capsules are grouped by different types, the learning of affine transformation ability between different types is reduced. Therefore, for the dataset with high affine transformation capability, the results of new network architecture are not as good as the original capsule network architecture.

 For the new $c_{ij}$ initialization method(The new network architecture $c_{ij}$ is initialized to approximately $0.028$ and the original capsule network architecture $c_{ij}$ is initialized to approximately $0.0008$), the experimental results of the new capsule network(Algorithm\ref{routingalg4}) is OurNet-c and the original capsule networks(Algorithm\ref{routingalg2}) is CapsNet-c are shown in {\bf Table \ref{tab3}}. 
\begin{table*}[h]
\caption{\label{tab3}Experimental results of Algorithm\ref{routingalg1}, Algorithm\ref{routingalg2} and Algorithm\ref{routingalg4}}
\centering
\begin{tabular}{|l|l|l|l|l|l|l|l|}
\hline
Model & MNIST & F-MNIST & K-MNIST & SVHN & CIFAR-10 & SMALLNORB & AFFNIST\\
\hline
CapsNet & 99.65 & 93.02 & 98.30 & 93.65 & 76.05 & 89.70 &99.66\\
\hline
CapsNet-c & 99.61 & 92.45 & 97.44 & 91.94 & 72.82 & 82.71 & 99.51\\
\hline
OurNet-c &  \bf 99.67 & \bf 93.13 & \bf 98.32 & \bf 93.81 & \bf 76.17 & \bf 87.49& \bf 99.59\\
\hline
\end{tabular}
\end{table*}

From the {\bf Table \ref{tab3}}, if the $c_{ij}$ is a new initialization method, the experimental results of the new network architecture are better than the original capsule network. At the same time, it is better than the original capsule network using the original $c_{ij}$ initialization method.

\begin{figure}[H]
\begin{center}
\includegraphics[width=1\textwidth]{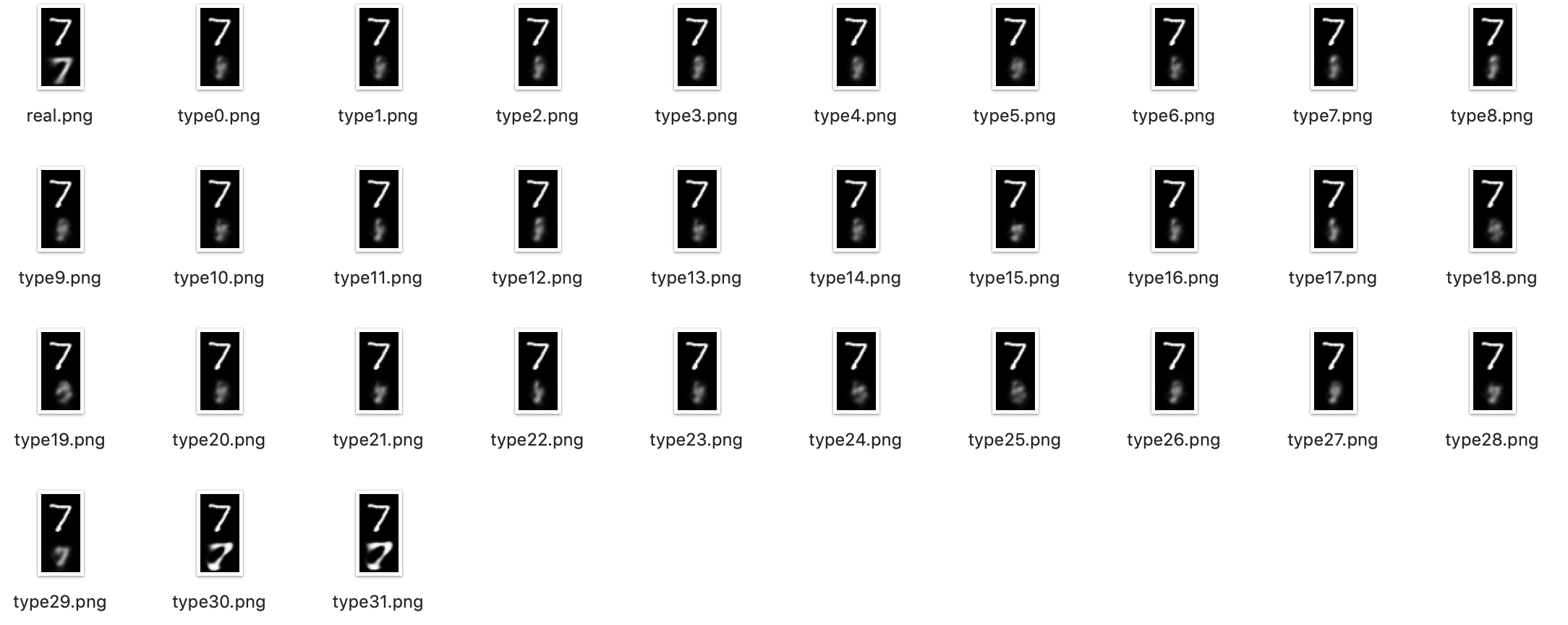}
\caption{Reconstructed images of different types of digit capsules}\label{p4}
\end{center}
\end{figure}

\begin{figure}[htb]
\begin{center}
\includegraphics[width=1\textwidth]{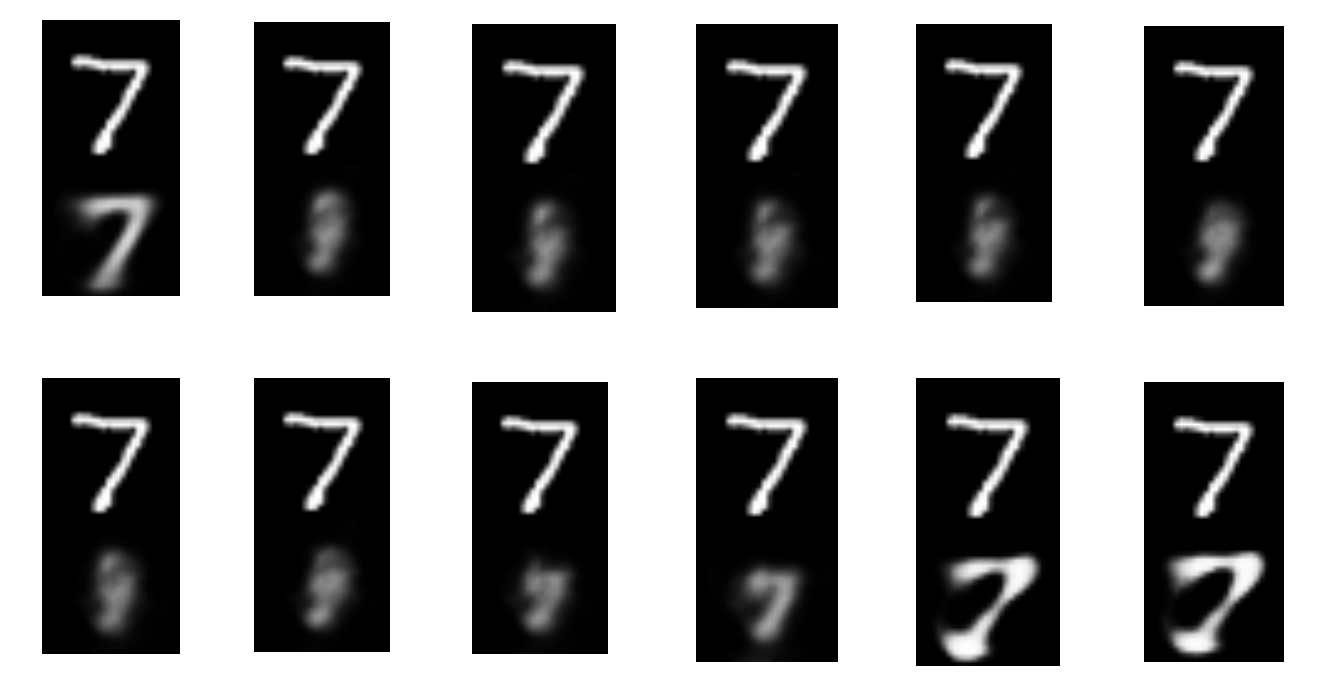}
\caption{Some Obvious Reconstructed images From Figure \ref{p4}}\label{p3}
\end{center}
\end{figure}

The Figure \ref{p4} reconstructed images of different types of capsules. The {\bf real.png} is the reconstruction of final digit capsule and others are different types of capsules. We select some representative reconstruction in Figure \ref{p3}.

Form the Figure \ref{p3}, We get the final digit capsule reconstructed image is smoother than the original image, better robust, and has the ability to denoise. For different types of reconstruction, The last two reconstruction look more like the original image and has distinct edges and corners. It shows the more similar the capsule type to the final digit capsule, the more obvious its characteristics. 

\begin{figure}[H]
\begin{minipage}[H]{0.5\linewidth}
\centering
\includegraphics[width=2.5in]{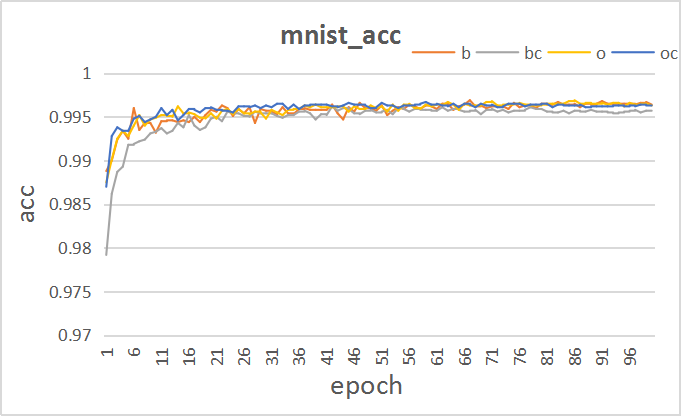}
\caption{mnist-acc}
\label{mnist-acc}
\end{minipage}%
\begin{minipage}[H]{0.5\linewidth}
\centering
\includegraphics[width=2.5in]{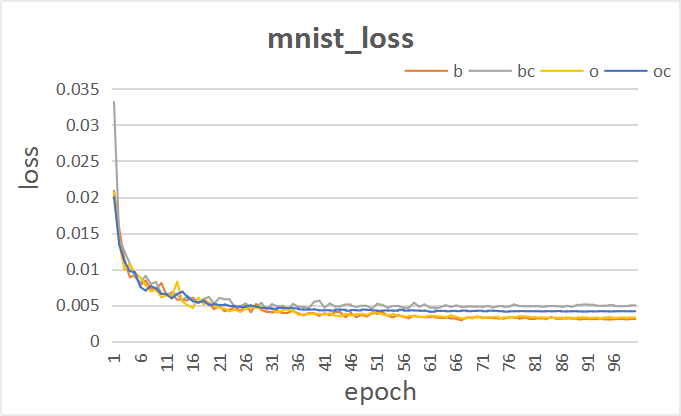}
\caption{mnist-loss}
\label{mnist-loss}
\end{minipage}
\end{figure}

\begin{figure}[H]
\begin{minipage}[H]{0.5\linewidth}
\centering
\includegraphics[width=2.5in]{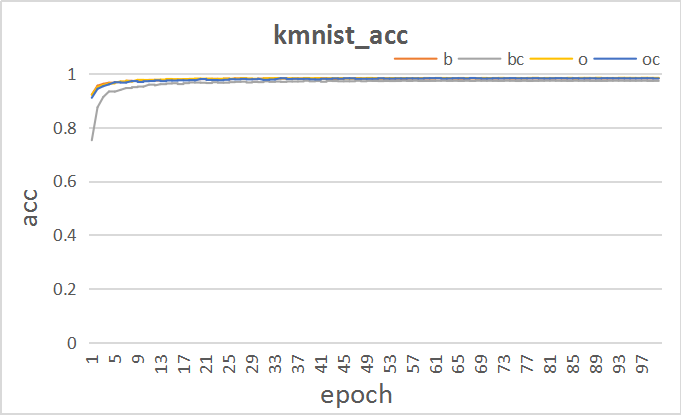}
\caption{kmnist-acc}
\label{kmnist-acc}
\end{minipage}%
\begin{minipage}[H]{0.5\linewidth}
\centering
\includegraphics[width=2.5in]{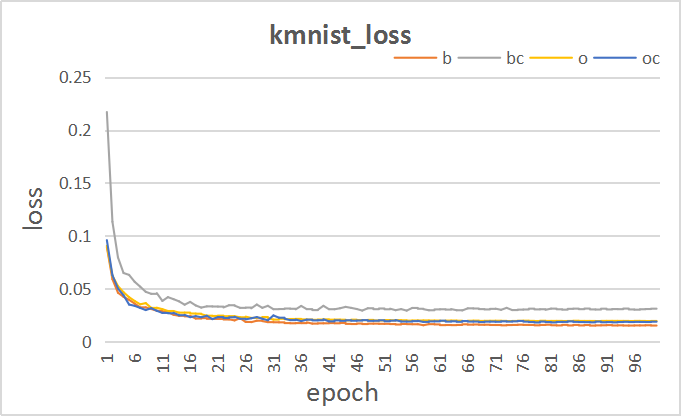}
\caption{kmnist-loss}
\label{kmnist-loss}
\end{minipage}
\end{figure}

\begin{figure}[H]
\begin{minipage}[H]{0.5\linewidth}
\centering
\includegraphics[width=2.5in]{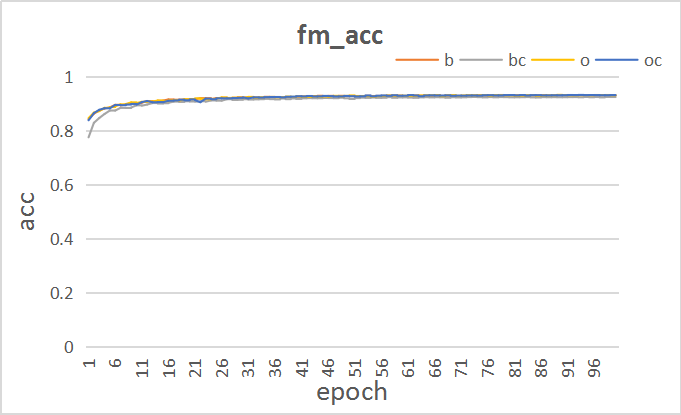}
\caption{fmnist-acc}
\label{fm-acc}
\end{minipage}%
\begin{minipage}[H]{0.5\linewidth}
\centering
\includegraphics[width=2.5in]{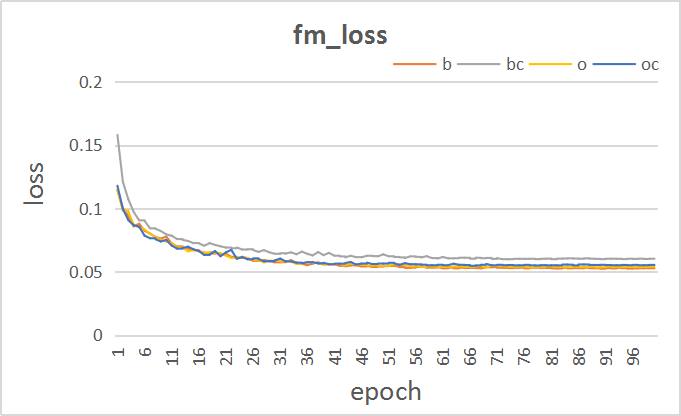}
\caption{fmnist-loss}
\label{fm-loss}
\end{minipage}
\end{figure}

\begin{figure}[H]
\begin{minipage}[H]{0.5\linewidth}
\centering
\includegraphics[width=2.5in]{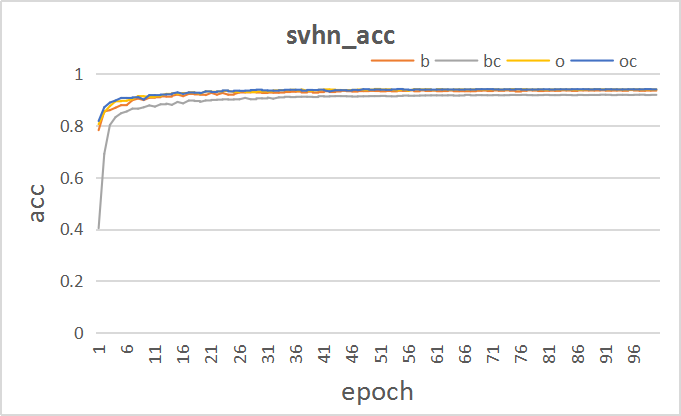}
\caption{svhn-acc}
\label{svhn-acc}
\end{minipage}%
\begin{minipage}[H]{0.5\linewidth}
\centering
\includegraphics[width=2.5in]{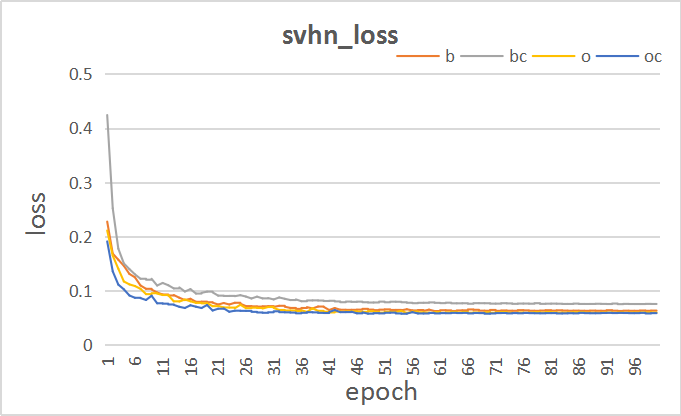}
\caption{svhn-loss}
\label{svhn-loss}
\end{minipage}
\end{figure}

\begin{figure}[H]
\begin{minipage}[H]{0.5\linewidth}
\centering
\includegraphics[width=2.5in]{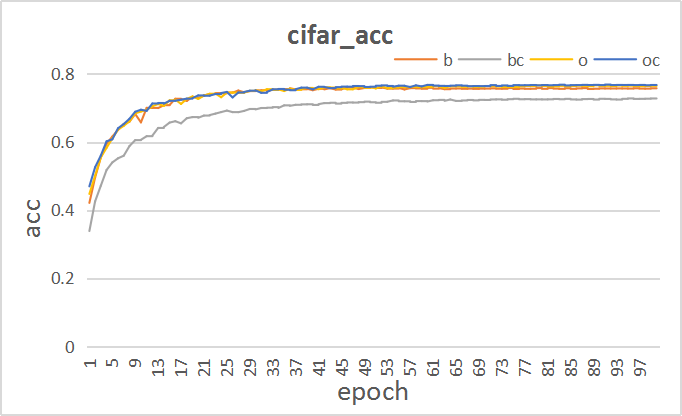}
\caption{cifar-acc}
\label{cifar-acc}
\end{minipage}%
\begin{minipage}[H]{0.5\linewidth}
\centering
\includegraphics[width=2.5in]{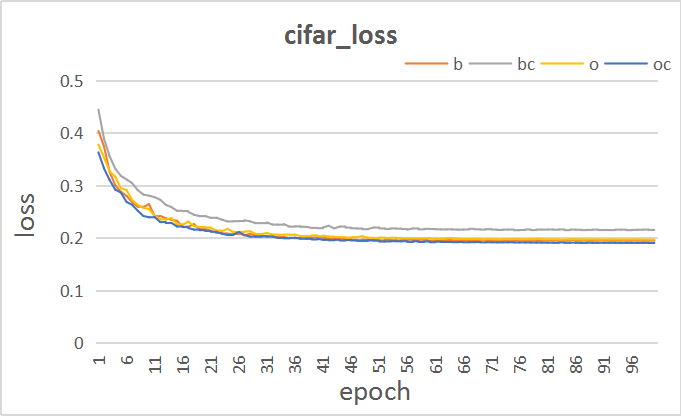}
\caption{cifar-loss}
\label{cifar-loss}
\end{minipage}
\end{figure}

The Figure \ref{mnist-acc}, Figure \ref{mnist-loss}, Figure \ref{kmnist-acc}, Figure \ref{kmnist-loss}, Figure \ref{fm-acc}, Figure \ref{fm-loss}, Figure \ref{svhn-acc}, Figure \ref{svhn-loss}, Figure \ref{cifar-acc} and Figure \ref{cifar-loss} shows the trends of accuracy and loss during the training of five datasets MNIST, K-MNIST, F-MNIST, SVHN and CIFAR-10. Lines $b$ and line $o$ in the Figure represent the experimental results of original capsule network and our improved capsule network, respectively. Lines $bc$ and line $oc$ in the Figure represent the experimental results of original capsule network and our improved capsule network with initialize the value of $c_{ij}$ using Algorithm \ref{routingalg2} and Algorithm \ref{routingalg4}, respectively.


\section{Conclusion and Future work}
This paper we proposed a novel capsule network architecture called the Grouped Capsules Networks(G-CapsNet). First, we discuss the initialization method of the coupling coefficient $c_{ij}$ in dynamic routing in the original capsule network. We consider ${\mathbf{s}}_j$ is the weighted sum of all of its corresponding input capsules in the lower layer and $c_{ij}$ is the softmax of all the capsules in the layer below calculated by $b_{ij}$ is more reasonable. But its experimental results are not good. Then we analyzed the value of $c_{ij}$ initialization and proposed to train different types of capsules. This improvement can adjust the value of $c_{ij}$ initialization to make it more suitable. Finally, we experimented on some computer vision datasets, the experimental results show that our improvement is effective and has achieved better results than the original capsule network.

Future work we will continue to research the algorithm improvement and model improvement of the capsule networks, and conduct in-depth research on the direction of algorithm optimization and model optimization.


\bibliographystyle{plain}
\bibliography{acml19}





\end{document}